%% file: paper.tex
\documentclass{article}
\usepackage{microtype}
\usepackage{graphicx}
\usepackage{adjustbox}
\usepackage{booktabs} 
\usepackage{hyperref}

\usepackage[accepted]{arxiv}
\usepackage{etex}
\usepackage[utf8]{inputenc} 
\usepackage[T1]{fontenc}    
\usepackage{hyperref}       
\usepackage{url}            
\usepackage{booktabs}       
\usepackage{amsfonts}       
\usepackage{nicefrac}       
\usepackage{microtype}      
\usepackage{bm}
\usepackage[mathscr]{eucal}
\usepackage{epstopdf}
\usepackage{makecell}
\usepackage{multirow}
\usepackage{graphicx}
\usepackage{wrapfig}
\usepackage{subcaption}
\usepackage{javen}
\usepackage{times}
\usepackage{epsfig}
\usepackage{amsmath}
\usepackage{amssymb}
\usepackage{adjustbox}
\usepackage{caption}
\usepackage{todonotes}

\newcommand{\cg}{\mathcal{G}}
\newcommand{\m}{\mathcal{M}}
\newcommand{\cd}{\mathcal{D}}
\newcommand{\dataset}{\mathbb{D}}
\newcommand{\train}{\mathrm{train}(\mathcal{T})}
\newcommand{\traini}{\mathrm{train}(\mathcal{T}_i)}
\newcommand{\testi}{\mathrm{test}(\mathcal{T}_i)}
\newcommand{\test}{\mathrm{test}(\mathcal{T})}
\newcommand{\learner}{f}
\newcommand{\task}{\mathcal{T}}
\newcommand{\taski}{\mathcal{T}_i}
\newcommand{\name}{\text{DEML}}
\newcommand{\commentx}[1]{{\iffalse #1 \fi}}
\DeclareMathAlphabet{\mathcal}{OMS}{cmsy}{m}{n}

\begin{document}
	\twocolumn[
    \vskip -0.3in
	\arxivtitle{Deep Meta-Learning: Learning to Learn in the Concept Space}
	
    \begin{arxivauthorlist}
        \vskip -0.1in
        \arxivauthor{Fengwei Zhou}{}\hspace{0.8cm}
        \arxivauthor{Bin Wu}{}\hspace{0.8cm}
        \arxivauthor{Zhenguo Li}{}\\
        Huawei Noah's Ark Lab\\
        \texttt{\{zhou.fengwei, wu.bin1, li.zhenguo\}@huawei.com}
    \end{arxivauthorlist}
	
	\vskip 0.2in
	]

	\input{abstract}

	\input{introduction}

\input{related}
	\input{model}

	\input{experiment}

	\input{conclusion}

	{\small
		\setlength{\bibsep}{5pt}
		\bibliography{meta-learning}
		\bibliographystyle{arxiv}
	}

\end{document}

%% file: abstract.tex
\begin{abstract}
Few-shot learning remains challenging for meta-learning that learns a learning algorithm (meta-learner) from many related tasks. In this work, we argue that this is due to the lack of a good representation for meta-learning,
and propose deep meta-learning to integrate the representation power of deep learning into meta-learning. The framework is composed of three modules, a concept generator, a meta-learner, and a concept discriminator, which are learned jointly. The concept generator, e.g. a deep residual net, extracts a representation for each instance that captures its high-level concept, on which the meta-learner performs few-shot learning, and the concept discriminator recognizes the concepts. By learning to learn in the concept space rather than in the complicated instance space, deep meta-learning can substantially improve vanilla meta-learning, which is demonstrated on various few-shot image recognition problems. For example, on 5-way-1-shot image recognition on CIFAR-100 and CUB-200, it improves Matching Nets from 50.53\% and 56.53\% to 58.18\% and 63.47\%, improves MAML from 49.28\% and 50.45\% to 56.65\% and 64.63\%, and improves Meta-SGD from 53.83\% and 53.34\% to 61.62\% and 66.95\%, respectively.

\end{abstract}

%% file: introduction.tex
\section{Introduction}
\begin{figure*}[ht!]
	\hspace{-0.0cm}\centering
	\includegraphics[scale=0.5]{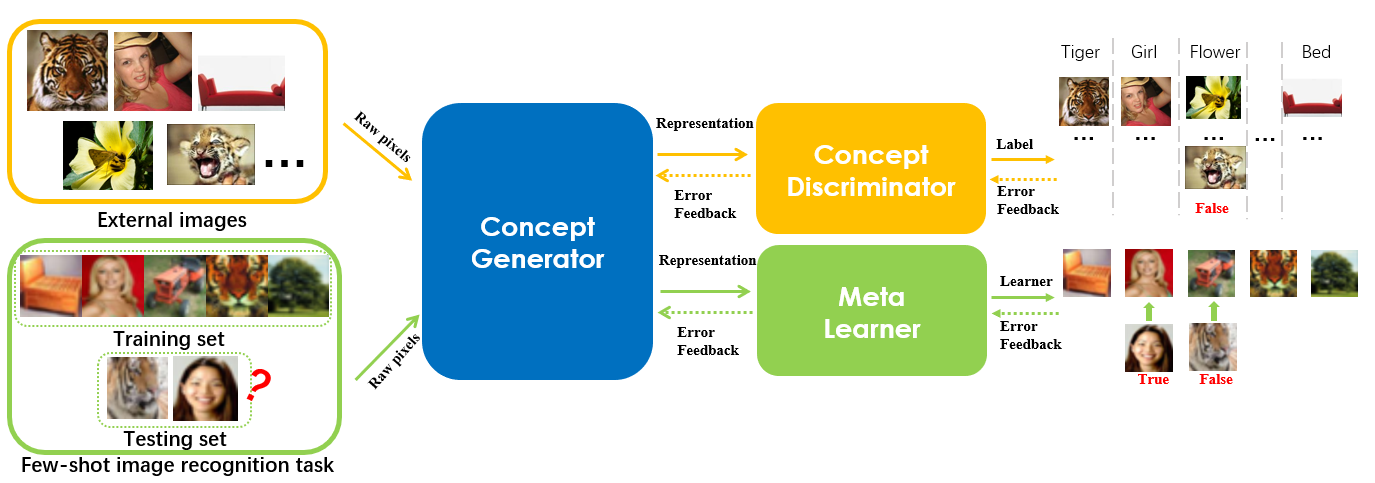}
    \vspace{-0.15cm}
	\caption{Deep meta-learning: learning to learn in the concept space. The concept generator learns to extract concept-level representations to ease meta-learning, while being enhanced by the concept discriminator that recognizes the concepts.
	}
	\label{fig:structure}
	\vspace{-0.5cm}
\end{figure*}

Many successes of machine learning today rely on enormous amounts of labeled data, which is not practical for problems with small data. For new applications such as autonomous vehicles, it is crucial to adapt instantly in a dynamic environment. In either cases, learning algorithms are required to consume labeled data efficiently. As one remarkable example, humans can learn new concepts rapidly from single images by leveraging knowledge learned before~\cite{lake2015human}.
However, most learning algorithms, especially those of deep learning, are data-hungry and do not function otherwise. Recently, meta-learning, pioneered by \cite{schmidhuber1987evolutionary}, draws renewed interest which learns on the level of tasks instead of instances, and learns task-agnostic learning algorithms (e.g. SGD) instead of task-specific models (e.g. CNN).
Remarkably, once trained, it can learn new tasks quickly from only a few examples (few-shot learning).

In meta-learning, one learns from a set of ``labeled'' tasks, as opposed to labeled instances, where each task consists of a training set and a test set. The goal is to fit to the tasks a learning algorithm that generalizes well to related new tasks, i.e., it can learn from the training data a learner (a model) that performs well on the test data. It involves learning at two levels -- gradual learning performed across tasks to gain meta-level knowledge and rapid learning carried out for a new task which is guided by the knowledge learned before.
Meta-learning has been shown to significantly outperform conventional learning on various few-shot learning problems, ranging from classification~\cite{santoro2016meta,vinyals2016matching,ravi2017optimization,finn2017model,li2017meta,snell2017prototypical}, reinforcement learning~\cite{wang2016learning,Yan2016reinf,mishra2018a,frans2018meta}, regression~\cite{santoro2016meta,finn2017model}, machine translation~\cite{kaiser2017learning}, to object tracking~\cite{park2018meta}. However, its full potential is still far from reach. For example, the accuracy of 5-way 5-shot recognition of natural images is around 60\% despite humans surpass 90\% with ease. We argue that this is due to the lack of a good data representation for meta-learning.

Few-shot learning is inherently challenging in the complicated instance space, where a few examples are insufficient to describe the intended high-level information such as categories or concepts. Consider image recognition, where each class corresponds to an abstract concept of objects such as ``cat'' or ``dog''. As an object is subject to a number of variations in scale, pose, translation, occlusion, illumination, distortion, background, etc,
the instance space can be highly complicated where two images of the same object can be drastically different in appearance. This makes few-shot learning almost intractable. While meta-learning alleviates the issue by leveraging many related tasks, it does not solve the problem. In this paper, we aim to show that  few-shot learning is much easier for meta-learning if done in the concept space.

Concepts more likely consist of rules rather than definitions~\cite{ahn1993psychological}. Instead of designing the rules by hand, we intend to leverage the power of deep learning. It is known that a deep convolutional neural network trained on large-scale image dataset can provide effective features for generic tasks~\cite{donahue2014decaf,razavian2014cnn,zeiler2014visualizing}. However, the meta-learner can hardly benefit from directly attaching the concept generator to it, especially when the previous image recognition tasks are quite different from the meta-learning tasks at hand. In our deep meta-learning framework, the key idea is to train a concept generator together with meta-learning tasks and large-scale image recognition tasks, which will finally improve the performance of vanilla meta learning methods. Specifically, a concept generator and a meta-learner are trained on a series of related tasks, and the concept generator is enhanced through a concept discriminator by handling image recognition tasks from an external dataset at the same time. This joint learning process can balance the concept learning from a large number of related few-shot image recognition tasks and from external large-scale datasets, which allows to incorporate both the external knowledge and task-agnostic meta-level knowledge into the concept generator. Also, this new meta-learning framework is a life-long learning system, where the concept generator can evolve continuously with the coming of fresh labeled samples.

Our main contributions can be summarized as follows:
\begin{itemize}
	\setlength\itemsep{0em}
  \item We propose deep meta-learning to integrate the power of deep learning into meta-learning, and show it improves vanilla meta-learning significantly on the problem of few-shot image recognition (see Figure~\ref{fig:structure}). We believe this framework is general and applicable to a variety of machine learning problems including reinforcement learning and regression.
  \item We propose to equip a meta-learner with a concept generator to enable learning to learn in the concept space while employing a concept discriminator to enhance the concept generator, and show that all three modules can be trained jointly in an end-to-end manner. Since the concept generator will continue to evolve with coming labeled data, this framework could literally be implemented as a life-long learning system.
  \item We instantiate the deep meta-learning framework on top of three state-of-the-art meta-learners including Matching Nets~\cite{vinyals2016matching}, MAML~\cite{finn2017model}, and Meta-SGD~\cite{li2017meta}, and conduct extensive experiments to show that deep meta-learning utilizes data more efficiently than all existing methods, and provides significantly better results on few-shot image recognition.
\end{itemize}

%% file: related.tex
\section{Related Work}
\textbf{Meta learning.} In~\cite{ravi2017optimization}, an LSTM is learned to train a learner such as CNN as it rolls out. \cite{finn2017model} learn how to initialize SGD while~\cite{li2017meta} learn a full-stack SGD, including initialization, update direction, and learning rate. In~\cite{vinyals2016matching}, a matching network with a non-parametric, differential KNN-like classifier is learned. In these methods, the meta-learner performs few-shot learning in the instance space, while in our proposed deep meta-learning, it is done in the concept space. Memory-augmented neural networks show high capacity for meta-learning. \cite{santoro2016meta} train an LSTM as a controller for accessing (read and write) an additional memory module, which is an extension of the internal memory in LSTM. The training process is time-consuming since the controller has to retrieve the entire memory at each time step. The memory is task-specific and is emptied once the task is finished. \cite{kaiser2017learning} also learn a matching network like~\cite{vinyals2016matching} but include a memory module that retains previously seen examples or their representatives. This method is designed to remember rare events, which is useful for machine translation. Like~\cite{santoro2016meta}, it is challenging to balance between the efficiency and accuracy of memory retrieval. Our concept generator can be considered as a memory module which memorizes the concept of each instance via a deep neural network, but it eliminates the need for exhaustive memory retrieval.

\textbf{Transfer learning and multi-task learning.} Transfer learning aims to transfer knowledge obtained from one domain with plenty of labeled data to another domain with few labeled data~\cite{pan2010survey}. Its performance depends on how relevant of the previous large-scale image recognition tasks to the tasks of interest~\cite{yosinski2014transferable}.In the deep learning regime, fine-tuning is a popular technique to perform transfer learning \cite{yosinski2014transferable}. However, the choices of \textit{frozen} layers and \textit{learning rate} should be manually tailored to avoid over-fitting and under-fitting. Also, this tedious labor work has to be done for every new task. In contrast, meta-learner is capable of rapidly adapting to new tasks automatically. Multi-task learning has been well studied in the literature, where the central idea is to jointly learn multiple related tasks via a shared representation~\cite{mtlbook}. For example, Fast R-CNN~\cite{girshick2015fast} trains image classifiers and bounding-box regressors simultaneously to perform object detection. Given 5-way-1-shot setting in a dataset of 100 classes, the number of 5-way classification tasks would incredibly reach to ${{100}\choose{5}} \times 5!$=9,034,502,400, and multi-task learning methods may fail in dealing with too many parameters induced by the large number of tasks ~\cite{argyriou2007multi,Ji:2009}. Meta-learning differs in that the meta-learner, once learned from many related tasks, can apply to any new task of the kind. Instead of measuring the similarity between tasks explicitly~\cite{Evgeniou:2005}, meta-learning methods could learn it more intrinsically.
\vspace{-0.1cm}

\textbf{Few-shot image recognition.} Quite a few methods have been proposed for few-shot image recognition.
\cite{FeifeiP2006} present a Bayesian model for learning categories from a few examples per category. \cite{pmlr-v27-salakhutdinov12a} organize seen categories into super-categories to derive hierarchically structured priors for new categories using a hierarchical Bayesian model. \cite{Lake_oneshot} develop a generative model that composes pen strokes into characters for handwritten character recognition. \cite{AWongICCV} extend this idea to natural images without relying on domain knowledge. Instead of following the typical gradient decent method, \cite{BertinettoHVTV16} suggest a feed-forward learner for learning deep neural networks to address the overfitting for few-shot learning. \cite{HariharanG16} generate dummy examples for the task of interest by using the geometric transformations inferred from existing examples of other categories. Recently, \cite{Xu_2017_CVPR} show a key-value memory networks for few-shot learning, which is not scalable due to the huge memory size and the heavy cost in memory retrieval. Besides, the fixed feature extractor makes it difficult to generalize to other domains. Other methods use metric learning to ease pairwise comparison between examples~\cite{fink2005object,koch2015siamese,isthatyou,schroff2015facenet}. Our method relies on meta-learning but learns to learn in the concept space.

%% file: model.tex
\section{Deep Meta-Learning}
In this section, we propose a new meta-learning framework, called deep meta-learning ($\name$), which integrates the representation power of deep learning into meta-learning, and enables learning to learn in the concept space.

\subsection{Framework}

Our framework~(Figure~\ref{fig:structure}) is composed of three modules, a concept generator $\cg$, a meta-learner $\m$, and a concept discriminator $\cd$, which are learned jointly. On one hand, we expect the concept generator $\cg$ extract task-agnostic meta-level representations that capture the high-level concepts of the instances from many related tasks, which can guide the meta-learner $\m$ to perform task-specific few-shot learning quickly \comment{for a new task with a few examples}. On the other hand, we hope that the concept generator $\cg$ can be enhanced through the concept discriminator $\cd$ by handling concept discrimination tasks on external large-scale datasets $\dataset$. After seeing a large number of instances and their concepts, the concept generator $\cg$ gradually learns the mapping from the raw instance space to the abstract concept space, and this high-capacity representation provider will greatly ease the meta-learning process.

Mathematically, we formulate the following optimization problem:
\begin{equation*}\label{eq:DEML_obj_short}
\small
\min_{\thetab_\cg, \thetab_\m, \thetab_\cd} \mathbb{E}_{\mathcal{T}\sim p(\mathcal{T}), (\xb,\yb)\sim \dataset}[J(\mathcal{L}_{\mathcal{T}}(\thetab_\m, \thetab_\cg), \mathcal{L}_{(\xb,\yb)}(\thetab_\cd, \thetab_\cg))],
\end{equation*}
where $\thetab_\cg, \thetab_\m$ and $\thetab_\cd$ are the parameters of corresponding modules. Meta-learning tasks $\task$ follow a distribution $p(\task)$ in a task space, and $(\xb, \yb)$ represents a labeled instance sampled from an external dataset $\dataset$. The objective is to minimize the expectation of the joint, denoted by $J$, of two losses: the loss $\mathcal{L}_{\mathcal{T}}(\thetab_\m, \thetab_\cg)$ on the meta-learning tasks and the loss $\mathcal{L}_{(\xb,\yb)}(\thetab_\cd, \thetab_\cg)$ on the concept discrimination tasks. The overall deep meta-learning process is summarized in Algorithm~\ref{algo:template}.
\begin{algorithm}[ht!]
	\caption{Deep Meta-Learning}
	\label{algo:template}
	\begin{algorithmic}
	\STATE {\bfseries Input: task distribution $p(\mathcal{T})$, labeled dataset $\dataset$, learning rate $\beta$
	}
	\STATE {\bfseries Output:$\thetab_\cg,\thetab_\cd,  \thetab_\m $}
	\STATE Initialize $\thetab_\cg,\thetab_\cd,  \thetab_\m$\;
	\WHILE {not done}
		\STATE Sample task batch $B_{t}$;
		\STATE Sample instance batch $B_i$;
		\STATE Compute meta-learning loss: $\mathcal{L}_{B_t}(\thetab_\m, \thetab_\cg)$;
		\STATE Compute concept discrimination loss: $\mathcal{L}_{B_i}(\thetab_\cd, \thetab_\cg)$;
		\STATE $(\thetab_\cg, \thetab_\cd,\thetab_{\m}) \leftarrow (\thetab_\cg, \thetab_\cd,\thetab_{\m})$\\ \rightline{$- \beta \nabla \Big[ J(\mathcal{L}_{B_t}(\thetab_\m, \thetab_\cg), \mathcal{L}_{B_i}(\thetab_\cd, \thetab_\cg)) \Big];$}
	\ENDWHILE
	\end{algorithmic}
\end{algorithm}

\subsection{Modules}

For the meta-learning pipeline $\m\circ\cg$, we assume that there is a distribution $p(\mathcal{T})$ over the related task space, from which we can sample tasks uniformly at random, and each task $\mathcal{T}$ consists of a training set $\train$ and a testing set $\test$. For the concept discrimination pipeline $\cd\circ\cg$, we assume that there is a large-scale labeled dataset $\dataset$ from which we can randomly sample labeled instances.

\textbf{Concept generator.} The concept generator $\cg$, parameterized by $\thetab_\cg$, is a deep neural network that could be any popular convolutional neural network such as AlexNet~\cite{krizhevsky2012imagenet}, Inception~\cite{szegedy2015going}, VGG~\cite{simonyan2014very}, or ResNet~\cite{he2016deep}.

\textbf{Concept discriminator.} The concept discriminator $\cd$, parameterized by $\thetab_\cd$,  is designed to predict labels for concepts generated by $\cg$. It could be implemented with any supervised learning method, such as support vector machines, nearest neighbor classifiers, or neural networks.

\textbf{Meta-learner.} The meta-learner $\m$\comment{(a learning algorithm)}, parameterized by $\thetab_\m$, learns to learn a learner for each task $\mathcal{T}$ based on $\train$. Any existing meta-learner can be used in our framework ~\cite{vinyals2016matching,ravi2017optimization,finn2017model,li2017meta,snell2017prototypical}. Matching Nets~\cite{vinyals2016matching} is a non-parametric model for few-shot learning based on metric learning. It learns from many related tasks a meta-learner (attention mechanism) which can guide the learner to do sample embedding. MAML~\cite{finn2017model} is another model designed for few-shot learning problems. It learns from many related tasks a meta-learner which can initialize a learner and update it by gradient descent with a fixed learning rate for each task $\task$ based on $\train$. Meta-SGD~\cite{li2017meta} suggests the meta-learner should learn not only the initialization, but also the update direction and learning rate of the learner.

In our deep meta-learning framework, the concept generator $\cg$, the concept discriminator $\cd$, and the meta-learner $\m$ are abstract modules which can be implemented using any proper models.

\subsection{Criterion}\label{sec:criterion}

The two pipelines, concept discrimination and meta-learning, are trained jointly in a learning-to-learn manner to optimize a combined loss. For the concept discrimination pipeline $\cd\circ\cg$, the concept generator is trained to generate representations for samples that capture their concepts and the concept discriminator is trained to discriminate the concepts. The goal is to minimize the expected loss on the concept discrimination tasks:
$$\mathcal{L}_{(\xb,\yb)}(\thetab_\cd, \thetab_\cg) = \ell(\cd\circ\cg(\xb), \yb),$$
where $\ell$ could be any loss function suitable for concept discrimination. For the meta-learning pipeline $\m\circ\cg$, the concept generator is trained to extract meta-level representations for samples and the meta-learner is learned to perform few-shot learning in the high-level concept space. Given the task $\task$, we define the meta-learner $\m$ as a mapping: $\m :\task \rightarrow \learner_{\task}$, i.e. $\m$ adapts a learner $\learner_{\task}$ for a task $\task$ through parametric~\cite{finn2017model,li2017meta} or non-parametric~\cite{vinyals2016matching} ways. The goal is to minimize the expected generalization loss on the meta-learning tasks:
$$\mathcal{L}_{\mathcal{T}}(\thetab_\m, \thetab_\cg) = \frac{1}{|\mathrm{test}(\mathcal{T})|}\sum_{(\xb,\yb)\in \mathrm{test}(\mathcal{T})}\ell(\learner_{\task}\circ\cg(\xb), \yb).$$
Its computation depends on the meta-learning algorithms selected. For example, if we choose Matching Nets~\cite{vinyals2016matching} as our meta-learner, then $\learner_{\task}\circ \cg(\xb)$ would be formalized as follows: 
$$\learner_{\task}\circ \cg(\xb) = \sum_{(\xb^\prime, \yb^\prime)\in \mathrm{train}(\mathcal{T})} a(\cg (\xb),\cg (\xb^\prime))\yb^\prime,$$
where
$$a(\cg (\xb),\cg (\xb^\prime)) = \frac{\mathrm{e}^{c(g( \cg (\xb)),g ( \cg (\xb^\prime)))}}{\sum_{(\xb^\prime, \yb^\prime)\in \mathrm{train}(\mathcal{T})} \mathrm{e}^{c(g (\cg (\xb)),g ( \cg (\xb^\prime)))}}$$
is the softmax over the cosine distance $c$ and the embedding function $g$\comment{parameterized by $\thetab_\m$}.

If MAML~\cite{finn2017model} is chosen as the meta-learner, then
$$\learner_{\task}\circ \cg(\xb) = \learner_{\thetab_{\m}-\alpha\nabla_{\thetab_{\m}}\mathcal{L}_{\train}(\thetab_{\m},\thetab_\cg)}(\cg(\xb))$$
where
$$\small\mathcal{L}_{\train}(\thetab_{\m},\thetab_\cg)=\frac{1}{|\mathrm{train}(\mathcal{T})|}\sum\limits_{(\xb, \yb) \in \mathrm{train}(\mathcal{T})} \ell(\learner_{\thetab_{\m}}(\cg(\xb)), \yb),$$
and $\alpha$ is a fixed learning rate.

Or if we choose Meta-SGD~\cite{li2017meta} as the meta-learner, then the parameter of $\m$ contains two parts $\thetab_{\m}=\{\phib,\alphab\}$ and $\m$ updates the initialization $\phib$ of the learner in the following way:
$$ \phib' = \phib- \alphab\circ\nabla_{\phib}\mathcal{L}_{\train}(\phib,\thetab_\cg).$$
The loss would be computed on $\test$ as follows:
$$\mathcal{L}_{\test}(\phib', \thetab_\cg) =
\frac{1}{|\mathrm{test}(\mathcal{T})|}\sum\limits_{(\xb,
	\yb)\in \mathrm{test}(\mathcal{T})} \ell(f_{\phib'}(\cg(\xb)), \yb).$$

\subsection{Algorithm}

After introducing the framework, modules, and criterion of deep meta-learning, we are ready to describe a complete algorithm of $\name$.
As a running case, we take Meta-SGD~\cite{li2017meta} as our meta-learner.
Recall that the meta-learning pipeline and the concept discrimination pipeline are trained synergistically to optimize a combined loss.
In this case, our deep meta-learning can be formulated as the following optimization problem:
\begin{multline*}\label{eq:DEML_SGD_obj}
\min_{\thetab_\cg, \thetab_\m=\{\phib, \alphab\}, \thetab_\cd} \mathbb{E}_{\mathcal{T}\sim p(\mathcal{T}), (\xb,\yb)\sim \dataset} [\mathcal{L}_{\mathrm{test}(\mathcal{T})}(\phib^\prime, \thetab_\cg)\\ + \lambda~\ell(\cd (\cg(\xb)), \yb)],
\end{multline*}
where $\phib' = \phib-\alphab\circ\nabla_{\phib}\mathcal{L}_{\mathrm{train}(\mathcal{T})}(\phib,\thetab_\cg)$ and $\lambda$ is a hyperparameter balancing meta-learning and concept discrimination.

The stochastic gradient descent (SGD) algorithm can be applied to optimize the above objective. In our implementation, we use the Adam~\cite{kingma2014adam} method, a variant of SGD. The detailed procedures are summarized in Algorithm~\ref{algo:train}, which is an instantiation of Algorithm~\ref{algo:template} by the use of Meta-SGD as the meta-learner.

\begin{algorithm*}[ht!]
	\caption{Deep Meta-Learning with Meta-SGD}
	\label{algo:train}
	\begin{algorithmic}[1]
		\STATE {\bfseries Input: task distribution $p(\mathcal{T})$, labeled dataset $\dataset$, batch size $n$ of tasks, batch size $m$ of instances, learning rate $\beta$
		}
		\STATE {\bfseries Output:$\thetab_\cg,\thetab_\cd,  \thetab_\m=\{\phib, \alphab\} $}
		\STATE Initialize $\thetab_\cg, \thetab_\cd,\phib, \alphab$\;
		\WHILE {not done}
		\STATE Sample $n$ tasks $\mathcal{T}_i\sim p(\mathcal{T})$ and $m$ instances $(\xb_j, \yb_j)\sim \dataset$\;
		\FOR{each $\taski$}
		\STATE \mbox{$\mathcal{L}_{\traini}(\phib, \thetab_\cg)\leftarrow \frac{1}{|\traini|} \sum\limits_{(\xb, \yb) \in \traini} \ell(\learner_{\phib} (\cg(\xb)), \yb)$;}
		\STATE $\phib^\prime_i \leftarrow \phib - \alphab \circ \nabla_{\phib} \mathcal{L}_{\traini}(\phib, \thetab_\cg)$;
		\STATE \mbox{$\mathcal{L}_{\testi}(\phib_i', \thetab_\cg) \leftarrow
			\frac{1}{|\mathrm{test}(\mathcal{T}_i)|}\sum\limits_{(\xb,
				\yb)\in \mathrm{test}(\mathcal{T}_i)} \ell(f_{\phib_i'}(\cg (\xb)), \yb)$;}
		
		\ENDFOR
		\STATE $(\thetab_\cg, \thetab_\cd,\phib, \alphab) \leftarrow (\thetab_\cg, \thetab_\cd,\phib, \alphab) - \beta \nabla \Big[\frac{1}{n}\sum\limits_{i=1}^n \mathcal{L}_{\testi}(\phib_i^\prime, \thetab_\cg)$
		{$+ \lambda~\frac{1}{m}\sum\limits_{j=1}^m \ell(\cd(\cg(\xb_j)), \yb_j)\Big]$;}
		\ENDWHILE
	\end{algorithmic}
	\vspace{-0.1cm}	
\end{algorithm*}

\subsection{Comparison with Related Work}
\textbf{Meta-learning.}
Vanilla meta-learning is done in the instance space, which can be challenging for the problem of few-shot learning because a few examples can hardly capture the high-level concept they represent (e.g. dog). In contrast, our framework executes meta-learning in the concept space thanks to the concept generator, where the problem of few-shot learning is much easier.

\textbf{Transfer learning.} A well-pretrained concept generator (e.g. neural network) could provide concept-level representation for a meta-learner, but it may benefit little if the novel tasks are quite different from those the concept generator is trained on.
Also, fine-tuning techniques may result in forgetting the concepts learned before.
In our deep meta-learning framework, we propose a principled way to train this concept generator to enhance the model, as well as a systematic way to use it to guide the few-shot learning tasks at hand in a learning-to-learn manner.

\textbf{Life-long learning.} Interestingly, this new deep meta-learning framework could be easily extended to a life-long learning system~\cite{silver2013lifelong}, which retains knowledge learned previously and adapts to new tasks over a lifelong time.
With the increase of labeled data for concept discrimination tasks, the concept generator could effectively retain learned concepts and provide more effective representations for the meta-learner. As a consequence, performance on new tasks will be improved gradually over time.

%% file: experiment.tex
\section{Experiments}

In this section, we evaluate the proposed deep meta-learning (DEML) on a number of few-shot image recognition problems, but note that it is applicable to classification, reinforcement learning, and regression in general.

\subsection{Datasets}
Two different pipelines in $\name$ process two different formats of data. For concept discrimination tasks, we perform experiments on a subset of ImageNet~\cite{imagenet_cvpr09}. For meta-learning tasks, we perform experiments on MiniImagenet~\cite{vinyals2016matching}, Caltech-256~\cite{griffin2007caltech}, CIFAR-100~\cite{Krizhevsky2009LearningML}, and CUB-200~\cite{WahCUB_200_2011}.

\subsubsection{Datasets for Concept Discrimination}
\textbf{ImageNet-200.} ImageNet\cite{imagenet_cvpr09} contains 14,197,122 images of 1,000 classes, including person, vehicle, plant, and so on. The whole dataset is too large, and so in our experiments we use a subset ImageNet-200 with 200 classes sampled from 900 classes (excluding the 100 classes used in MiniImagenet\cite{vinyals2016matching}). For images in each selected category, $90\%$ examples are randomly chosen for training, and the remaining images are used for testing.

\subsubsection{Datasets for Meta-Learning}
\noindent\textbf{MiniImagenet.} The MiniImagenet dataset, first used in~\cite{vinyals2016matching}, consists of 60,000 color images of 100 classes, with 600 examples per class. For our experiments, we use the splits introduced by~\cite{ravi2017optimization}. Their splits use a different set of 100 classes, which are divided into three disjoint subsets: 64 classes for meta-training, 16 classes for meta-validation, and 20 classes for meta-testing. Particularly, MiniImagenet and ImageNet-200 are mutually exclusive at class level.

\noindent\textbf{Caltech-256.} The Caltech-256 dataset~\cite{griffin2007caltech} is a successor to the well-known dataset Caltech-101. It consists of 30,607 color images of 256 classes. We use 150, 56, and 50 classes for meta-training, meta-validation, and meta-testing, respectively.

\noindent\textbf{CIFAR-100.} The CIFAR-100 dataset~\cite{Krizhevsky2009LearningML} contains 60,000 $32 \times 32$ color images of 100 classes. We use 64, 16, and 20 classes for meta-training, meta-validation, and meta-testing, respectively. In CIFAR-100, images are rescaled to $32 \times 32 $.
Consequently, the difficulty in recognizing different categories is greatly increased.

\noindent\textbf{Caltech-UCSD Birds-200-2011 (CUB-200).} The CUB-200 dataset~\cite{WahCUB_200_2011} contains 11,788 color images of 200 different bird species. We use 140 classes for meta-training, 20 classes for meta-validation, and test on the remaining 40 classes. In this fine-grained dataset, subtle differences between very similar classes can hardly be recognized even by humans.

\subsection{Baselines}\label{sec:baselines}
To compare $\name$ with existing meta-learning methods, we evaluate existing meta-learning methods (\textbf{Matching Nets}~\cite{vinyals2016matching}, \textbf{MAML}~\cite{finn2017model}, and \textbf{Meta-SGD}~\cite{li2017meta}) on meta-learning datasets as our baselines. We follow their original neural network designs to reproduce their results at first.
To show that the improvements of $\name$ \comment{compared to the previous approaches} are not solely attributed to the deeper neural network and rescaled images, we also evaluate the previous approaches with exactly the same architecture (excluding the concept discriminator) and inputs as $\name$. Accordingly, their deep version implementations are denoted by \textbf{Deep Matching Nets}, \textbf{Deep MAML}, and \textbf{Deep Meta-SGD}, respectively. Since $\name$ is a meta-learner-agnostic framework for meta-learning, we re-implement \textbf{Matching Nets}, \textbf{MAML} and \textbf{Meta-SGD} on \textbf{$\name$} with the following implementation configurations.
\begin{table*}[ht!]
	\centering

	\caption{Comparison between deep meta-learning and vanilla meta-learning.}
	\label{table:main}
	\begin{adjustbox}{max width=\textwidth}
		\begin{tabular}{@{}l|l|l|l|l|l|l|l|l@{}}
			\toprule
			\toprule
			\multirow{2}{*}{Method}&\multicolumn{2}{c|}{MiniImagenet}&\multicolumn{2}{c|}{Caltech-256}
			&\multicolumn{2}{c|}{CIFAR-100}&\multicolumn{2}{c}{CUB-200} \\
			\cmidrule{2-9}
			&5-way-1-shot&5-way-5-shot&5-way-1-shot&5-way-5-shot&5-way-1-shot&5-way-5-shot&5-way-1-shot&5-way-5-shot\\
			\midrule
			Matching Nets&43.56 $\pm$ 0.84&55.31 $\pm$ 0.73&48.09 $\pm$ 0.83&57.45 $\pm$ 0.74&50.53 $\pm$ 0.87&60.30 $\pm$ 0.82&56.53 $\pm$ 0.99&63.54 $\pm$ 0.85\\
			\midrule
			DEML+Matching Nets&55.84 $\pm$ 0.94&59.88 $\pm$ 0.73&52.97 $\pm$ 0.99&59.42 $\pm$ 0.75&58.18 $\pm$ 1.09&63.12 $\pm$ 0.85&63.47 $\pm$ 1.10&64.86 $\pm$ 0.87\\
			\midrule
			MAML&48.70 $\pm$ 1.84&63.11 $\pm$ 0.92&45.59 $\pm$ 0.77&54.61 $\pm$ 0.73&49.28 $\pm$ 0.90&58.30 $\pm$ 0.80&50.45 $\pm$ 0.97&59.60 $\pm$ 0.84\\
			\midrule
			DEML+MAML&53.71 $\pm$ 0.89&68.13 $\pm$ 0.77&56.81 $\pm$ 1.01&70.54 $\pm$ 0.73&56.65 $\pm$ 1.09&68.66 $\pm$ 0.85&64.63 $\pm$ 1.08&66.75 $\pm$ 0.89\\
			\midrule
			Meta-SGD&50.47 $\pm$ 1.87&64.03 $\pm$ 0.94&48.65 $\pm$ 0.82&64.74 $\pm$ 0.75&53.83 $\pm$ 0.89&70.40 $\pm$ 0.74&53.34 $\pm$ 0.97&67.59 $\pm$ 0.82\\
			\midrule
			DEML+Meta-SGD&\textbf{58.49 $\pm$ 0.91}&\textbf{71.28 $\pm$ 0.69}&\textbf{62.25 $\pm$ 1.00}&\textbf{79.52 $\pm$ 0.63}&\textbf{61.62 $\pm$ 1.01}&\textbf{77.94 $\pm$ 0.74}&\textbf{66.95 $\pm$ 1.06}&\textbf{77.11 $\pm$ 0.78}\\
			\bottomrule
			\bottomrule
		\end{tabular}
	\end{adjustbox}
\vspace{-0.4cm}
\end{table*}

\subsection{Implementation}
\begin{figure}[t]
	\centering
	\includegraphics[scale=0.28]{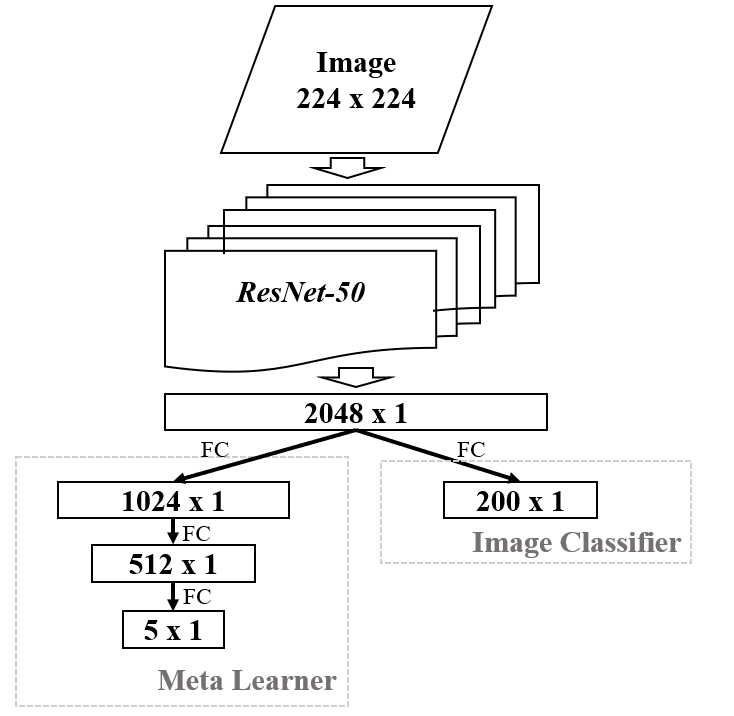}
	\caption{Model configuration in DEML for few-shot image recognition.}
	\label{fig:model}
	\vspace{-0.7cm}
\end{figure}

According to the analysis of~\cite{canziani2016analysis}, we choose ResNet-50~\cite{he2016deep} (excluding the last layer) as our concept generator in the experiments. For the concept discriminator, we use a shallow neural network with one fully connected layer. The learner for few-shot learning tasks depends on the meta-learner we choose. When choosing Matching Nets as the meta-learner, the learner is a neural network with an input layer of size 2048, followed by one hidden layer of size 1024 with ReLU nonlinearities, and then an output layer of size 512. When choosing MAML or Meta-SGD as the meta-learner, the learner is a neural network with the same input layer, followed by two hidden layers of size 1024 and 512 with ReLU nonlinearities, and then an output layer of size 5. The architecture of our deep meta-learning is provided in Figure~\ref{fig:model}\footnote{Images with different input sizes will be rescaled to 224x224.}. After introducing the architecture of $\name$ in our experiments, more experiment design details for meta-training and meta-testing are provided below.

\textbf{Meta-training.} For each iteration, a batch of examples from ImageNet-200 is sampled for the image recognition pipeline $\cd\circ\cg$. The prediction loss is measured by the mean of cross-entropy over all the examples in this batch. For the meta-learning pipeline $\m\circ\cg$, a batch of tasks is sampled from one specific \comment{of the} meta-learning dataset. Here, each task contains 5 classes of examples, each with $K\in\{1,5\}$ examples for training and 5 examples for testing. The meta-learner will generate one learner based on the training set of each task, and then is evaluated on the testing set, as stated in Section~\ref{sec:criterion}. The prediction loss is also measured by the mean of cross-entropy over all the examples in the testing sets of tasks in the batch. We update the whole model once each iteration finishes according to Algorithm~\ref{algo:train}. The parameter $\lambda$ is set to 1.0 in our experiments at first, and further studies on different values will also be discussed later. The batch size of examples for image recognition is set to 64 and the batch size of tasks is set to 4 and 2 for 1-shot and 5-shot recognition, respectively. The number of iterations is 60,000 in the experiments on MiniImagenet, Caltech-256, and CIFAR-100, and 20,000 in the experiments on CUB-200.

\textbf{Meta-testing.} For performance evaluation, we randomly sample 600 tasks from the corresponding meta-learning dataset. Each task contains 5 classes of examples, each with $K\in\{1,5\}$ examples for training and 15 examples for testing. The results averaged over the sampled 600 tasks with $95\%$ confidence intervals are reported at Section \ref{sec:result}.

For both MAML and Meta-SGD, the meta-learner uses one-step adaptation during meta-training and meta-testing for fair, and the learning rate $\alpha$ for MAML is set to 0.01 in all experiments.

\subsection{Results and Discussion}\label{sec:result}
\textbf{$\name$ version vs.\ vanilla version.} The comparison results between $\name$ versions and vanilla versions of Matching Nets, MAML, and Meta-SGD are summarized in Table \ref{table:main}.

Our results clearly indicate that $\name$ versions outperform vanilla versions by a wide margin. The effective representations provided by the concept generator can ease the meta-learning process for different kinds of meta-learners. Our framework lifts the meta-learning from the complicated instance space to the high-level concept space and achieves high accuracy.

\textbf{$\name$ version vs.\ vanilla deep version.} To validate that the improvements of $\name$ are not merely because of the deeper neural network and rescaled images, we also evaluate the deep versions of the previous approaches on MiniImagenet as mentioned in Section~\ref{sec:baselines}. We enlarge the meta-training dataset by merging together the original 64 classes of MiniImagenet and the 200 classes of ImageNet-200. The results are summarized in Table~\ref{table:large_nets}. It can be seen that simply enlarging the network and training dataset can not lead to a higher accuracy. $\name$ leverages the power of deep learning in a more principled way and achieves superior performance.

\begin{table}[t]
	\centering
	\caption{Comparison between deep meta-learning and vanilla meta-learning (deep version).}
	\label{table:large_nets}
	\begin{tabular}{l|l|l}
		\toprule
		\toprule
		\multirow{2}{*}{Method}&\multicolumn{2}{c}{MiniImagenet}\\
		\cmidrule{2-3}
		&5-way-1-shot&5-way-5-shot\\
		\midrule
		Deep Matching Nets&48.82 $\pm$ 0.89&55.22 $\pm$ 0.70\\
		\midrule
		DEML+Matching Nets&55.84 $\pm$ 0.94&59.88 $\pm$ 0.73\\
		\midrule
		Deep MAML&51.74 $\pm$ 0.94&57.24 $\pm$ 0.74\\
		\midrule
		DEML+MAML&53.71 $\pm$ 0.89&68.13 $\pm$ 0.77\\
		\midrule
		Deep Meta-SGD&51.62 $\pm$ 0.95&64.50 $\pm$ 0.74\\
		\midrule
		DEML+Meta-SGD&\textbf{58.49 $\pm$ 0.91}&\textbf{71.28 $\pm$ 0.69}\\
		\bottomrule
		\bottomrule
	\end{tabular}
\vspace{-0.5cm}
\end{table}

\begin{table*}[ht!]
	\centering
	{
		
		\caption{Comparison between deep meta-learning and transfer learning.}
		\label{table:transfer}
		\begin{adjustbox}{max width=\textwidth}
			\begin{tabular}{@{}l|l|l|l|l|l|l|l|l@{}}
				\toprule
				\toprule
				\multirow{2}{*}{Method}&\multicolumn{2}{c|}{MiniImagenet}&\multicolumn{2}{c|}{Caltech-256}
				&\multicolumn{2}{c|}{CIFAR-100}&\multicolumn{2}{c}{CUB-200} \\
				\cmidrule{2-9}
				&5-way-1-shot&5-way-5-shot&5-way-1-shot&5-way-5-shot&5-way-1-shot&5-way-5-shot&5-way-1-shot&5-way-5-shot\\
				\midrule
				Decaf+kNN&\textbf{61.81 $\pm$ 0.84}&\textbf{79.88 $\pm$ 0.58}&\textbf{63.08 $\pm$ 0.89}&\textbf{80.70 $\pm$ 0.61}&47.04 $\pm$ 0.80&65.96 $\pm$ 0.73&45.58 $\pm$ 0.78&65.57 $\pm$ 0.70\\
				\midrule
				Decaf+Meta-SGD&58.06 $\pm$ 0.86&71.30 $\pm$ 0.70&60.47 $\pm$ 0.92&74.91 $\pm$ 0.70&54.10 $\pm$ 0.98&68.30 $\pm$ 0.73&56.85 $\pm$ 0.98&66.29 $\pm$ 0.78\\
				\midrule
				DEML+Meta-SGD&58.49 $\pm$ 0.91&71.28 $\pm$ 0.69&62.25 $\pm$ 1.00&79.52 $\pm$ 0.63&\textbf{61.62 $\pm$ 1.01}&\textbf{77.94 $\pm$ 0.74}&\textbf{66.95 $\pm$ 1.06}&\textbf{77.11 $\pm$ 0.78}\\
				\bottomrule
				\bottomrule
			\end{tabular}
		\end{adjustbox}
		\vspace{-0.4cm}
	}
\end{table*}
\textbf{$\name$ vs.\ transfer learning.} To compare deep meta-learning with transfer learning, we also evaluate some variants of $\name$. One simple baseline is \textbf{Decaf+kNN}, where we merely train our concept generator and discriminator $\cd\circ\cg$ on ImageNet-200 with 60000 episodes. For each task at meta-testing time, we compute a centroid for each class by averaging the features of the training examples, produced by the concept generator $\cg$,\comment{of training examples of this class,} and label each testing example with its nearest (Euclidean distance) centroid's category\comment{then use nearest-neighbor matching (Euclidean distance) among the centroid to classify the testing examples}.  Another simplified version of $\name$ is \textbf{Decaf+Meta-SGD},
where one pretrained generator $\cg$ is attached to the meta-learner to execute meta-training process. At meta-training time, we exclude the image recognition pipeline and fix the parameters of the concept generator. The results are shown in Table~\ref{table:transfer}.

It is interesting to note that the baseline Decaf+kNN achieves the best performance on MiniImagenet and Caltech-256. Since the concept generator $\cg$ is trained on ImageNet-200, which is quite similar to MiniImagenet and Caltech-256~\cite{tommasi2017deeper}, representations provided by it are so effective that the naive nearest-neighbor baseline achieves the best performance. This shows that the effective representations are quite crucial for few-shot learning and if we have some prior knowledge of the few-shot learning tasks, we can incorporate it into the concept generator by choosing particular concept recognition dataset to ease the meta-learning process. On the contrary, the performance of Decaf+kNN drops a lot on CIFAR-100 and CUB-200, since these two datasets are quite different from ImageNet-200. When the meta-learning dataset is quite different from the dataset where the concept generator is trained on, the meta-learner benefits little by directly adding this generator to it. In the joint learning process of $\name$, the concept generator extracts the task-agnostic meta-level concepts of the data, as well as the external concepts. Combining the two sources of knowledge, the concept generator provides effective representations for the meta-learner to do few-shot learning.

\begin{figure}[ht!]
	\centering
	\begin{subfigure}{0.23\textwidth}
		\centering
		\includegraphics[width=1.0\textwidth]{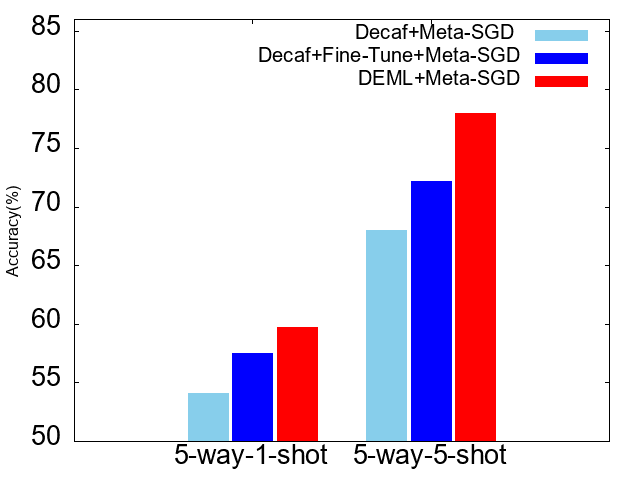}
	\end{subfigure}
	\begin{subfigure}{0.23\textwidth}
		\centering
		\includegraphics[width=1.0\textwidth]{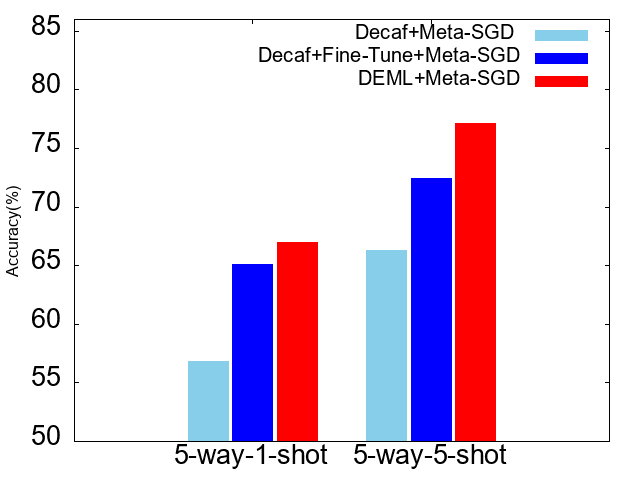}
	\end{subfigure}
	
	\caption{Comparison among Decaf+Meta-SGD, Decaf+Fine-Tune+Meta-SGD, and DEML+Meta-SGD on CIFAR-100 (left) and CUB-200 (right).}
	\label{fig:finetune}
	\vspace{-0.2cm}
\end{figure}

To emphasize the necessity of our joint learning process, we propose the third version \textbf{Decaf+Fine-Tune+Meta-SGD} which is the same as Decaf+Meta-SGD except that the concept generator and the meta-learner are trained together during meta-training process. The models are trained for 60,000 and 20,000 iterations on CIFAR-100 and CUB-200, respectively, and the results are shown in Figure~\ref{fig:finetune}, together with the results of Decaf+Meta-SGD and DEML+Meta-SGD. DEML+Meta-SGD performs consistently better than Decaf+Meta-SGD and Decaf+Fine-Tune+Meta-SGD on all cases by a wide margin. Our joint learning process can balance the learning from a large number of related few-shot learning tasks and the learning from external large-scale dataset. The concept generator being enhanced by the concept recognition pipeline as the meta-learning proceeds may lead to a higher generalization capability. The meta-learner and the concept generator evolve synergistically in this joint learning process.

\begin{figure}[ht!]
	\centering
	\includegraphics[width=0.8\linewidth]{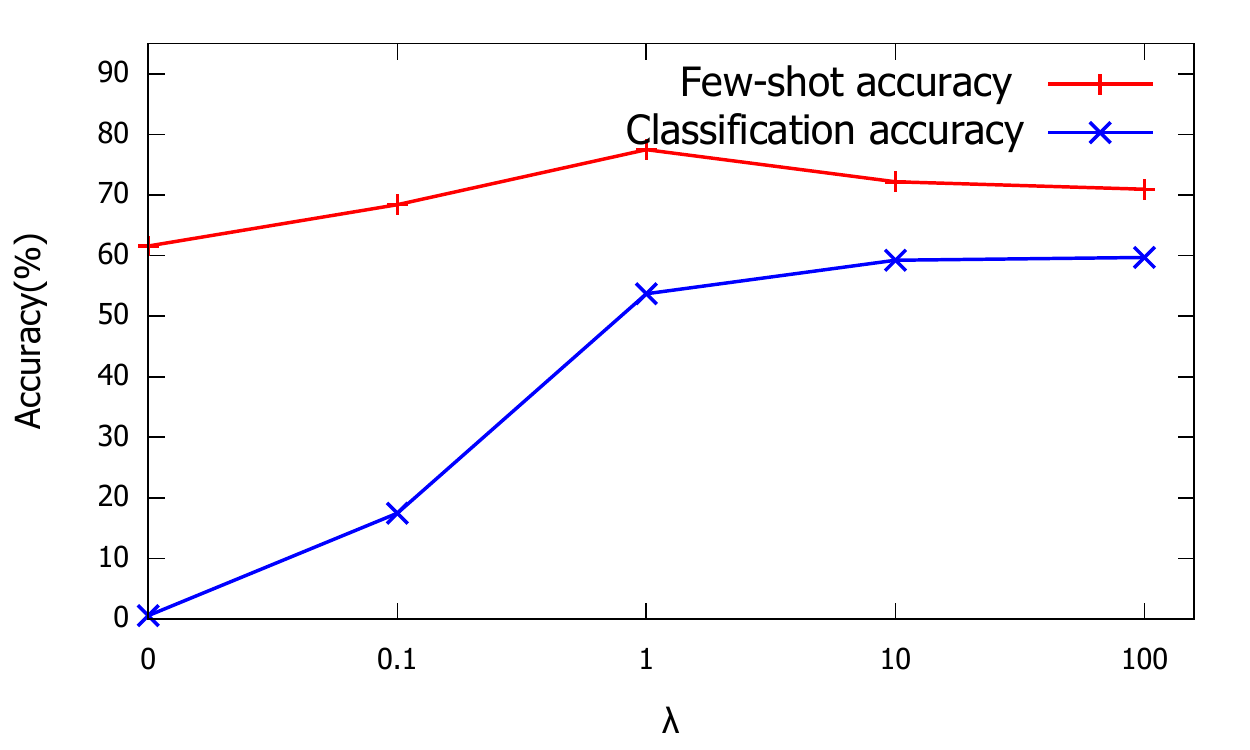}
	\caption{Few-shot (5-way-5-shot) learning accuracy and concept recognition accuracy of DEML+Meta-SGD on CIFAR-100 with different values of $\lambda$.}
	\label{fig:diff_lambda}
	\vspace{-0.6cm}
\end{figure}

\textbf{Study of $\lambda$.} In previous experiments, the hyperparameter $\lambda$ is set to 1.0 as default. For different datasets, one intuition is that we should incorporate external concepts at different levels. We verify it on CIFAR-100 with the 5-way-5-shot case with DEML+Meta-SGD (Figure~\ref{fig:diff_lambda}).
It is obvious that as the value of $\lambda$ increases, the accuracy of concept recognition increases accordingly. However, the accuracy of few-shot learning tasks increases first and then decreases. This result shows that the meta-learner does benefit from the concept generator enhanced by the external data, but placing too much emphasis on the external data can harm the performance of meta-learner. A balance between the external knowledge and the internal meta-level knowledge is useful in $\name$.

%% file: conclusion.tex
\section{Conclusion and Future Work}

In this paper, we propose deep meta-learning that integrates the representation power of deep learning into meta-learning, and enables learning to learn in the concept space. A concept generator that can provide effective representations for the meta-learner is trained on large-scale concept discrimination and few-shot learning, simultaneously. This high-capacity generator captures the concept information of examples from the external large-scale dataset as well as the meta-level concepts of data from a large number of related few-shot learning tasks, which lifts the meta-learning from the raw instance space to the high-level concept space and eases the meta-learning process. The joint learning process in this new framework allows the meta-learner and the concept generator evolve synergistically, which leads to a higher generalization capability. Extensive experiments on few-shot image recognition show that this new framework improves the vanilla meta-learning greatly.

In our experiments, we train the concept generator together with the concept discriminator on a single dataset with 200 classes. It would be interesting to train this concept generator on multiple large-scale datasets with a large number of categories, which can incorporate more external concepts into the model. This requires more careful design for the concept generator and more computing resources. We leave it for future work. Another future work is to implement a life-long learning system that can evolve with new examples and new concepts. Indeed, our algorithm enables life-long learning. The concept generator can be enhanced with more training examples coming. A balance between the concept learning from external datasets and from the few-shot learning tasks at hand is crucial in this life-long learning scenario. Forgetting problem is another concern. Learning new concepts should not result in the generator forgetting  previously learned concepts. More studies are expected to explore this problem.